\UseRawInputEncoding
\documentclass[journal]{IEEEtran}
\usepackage{amsmath,amsfonts}
\pdfoutput=1
\usepackage{array}
\usepackage[caption=false,font=normalsize,labelfont=sf,textfont=sf]{subfig}
\usepackage{textcomp}
\usepackage{stfloats}
\usepackage{url}
\usepackage{verbatim}
\usepackage{graphicx}
\usepackage{cite}
\usepackage{booktabs}
\usepackage{multirow}
\usepackage{hyperref}
\usepackage{float}
\usepackage{bbding}
\usepackage{bm}
\usepackage{threeparttable}   
\begin{document}
\title{CorrTalk: Correlation Between Hierarchical Speech and Facial Activity Variances for 3D Animation}

\author{Zhaojie Chu, Kailing Guo, Xiaofen Xing, Yilin Lan, Bolun Cai, and Xiangmin Xu, ~\IEEEmembership{Senior Member, IEEE}
\thanks{Zhaojie Chu, Kailing Guo, Xiaofen Xing, Yilin Lan, and Xiangmin Xu are with South China University of Technology, Guangzhou 510640, China. Bolun Cai is with ByteDance Inc. \textit{(Corresponding authors: Xiangmin Xu. )}}
}

\markboth{Journal of \LaTeX\ Class Files,~Vol.~14, No.~8, August~2021}%
{Shell \MakeLowercase{\textit{et al.}}: A Sample Article Using IEEEtran.cls for IEEE Journals}
\maketitle

\begin{abstract}
Speech-driven 3D facial animation is a challenging cross-modal task that has attracted growing research interest. During speaking activities, the mouth displays strong motions, while the other facial regions typically demonstrate comparatively weak activity levels. Existing approaches often simplify the process by directly mapping single-level speech features to the entire facial animation, which overlook the differences in facial activity intensity leading to overly smoothed facial movements. In this study, we propose a novel framework, CorrTalk, which effectively establishes the temporal correlation between hierarchical speech features and facial activities of different intensities across distinct regions. A novel facial activity intensity metric is defined to distinguish between strong and weak facial activity, obtained by computing the short-time Fourier transform of facial vertex displacements. Based on the variances in facial activity, we propose a dual-branch decoding framework to synchronously synthesize strong and weak facial activity, which guarantees wider intensity facial animation synthesis. Furthermore, a weighted hierarchical feature encoder is proposed to establish temporal correlation between hierarchical speech features and facial activity at different intensities, which ensures lip-sync and plausible facial expressions. Extensive qualitatively and quantitatively experiments as well as a user study indicate that our CorrTalk outperforms existing state-of-the-art methods. The source code and supplementary video are publicly available at: \url{https://zjchu.github.io/projects/CorrTalk/}.

\end{abstract}


\section{Introduction}
\IEEEPARstart3D digital humans have attracted substantial attention from academic communities and have found extensive applications in commercial communities such as computer games, virtual reality, and film production. Among these applications, a 3D avatar needs to be automatically driven by arbitrary input signals, such as speech or text, to obtain a vivid and realistic digital human. There are high dependencies between speech and facial gestures \cite{10143326,yu2020multimodal,aleksic2004speech,xuefeng,richard2021meshtalk}, and speech not only conveys much detailed content but also contains rich semantic context information\cite{liu2020mockingjay}. Thus, speech-driven 3D facial animation has sparked the interest of a growing number of researchers.

\begin{figure}
    \centering
    \includegraphics[width=3.4 in]{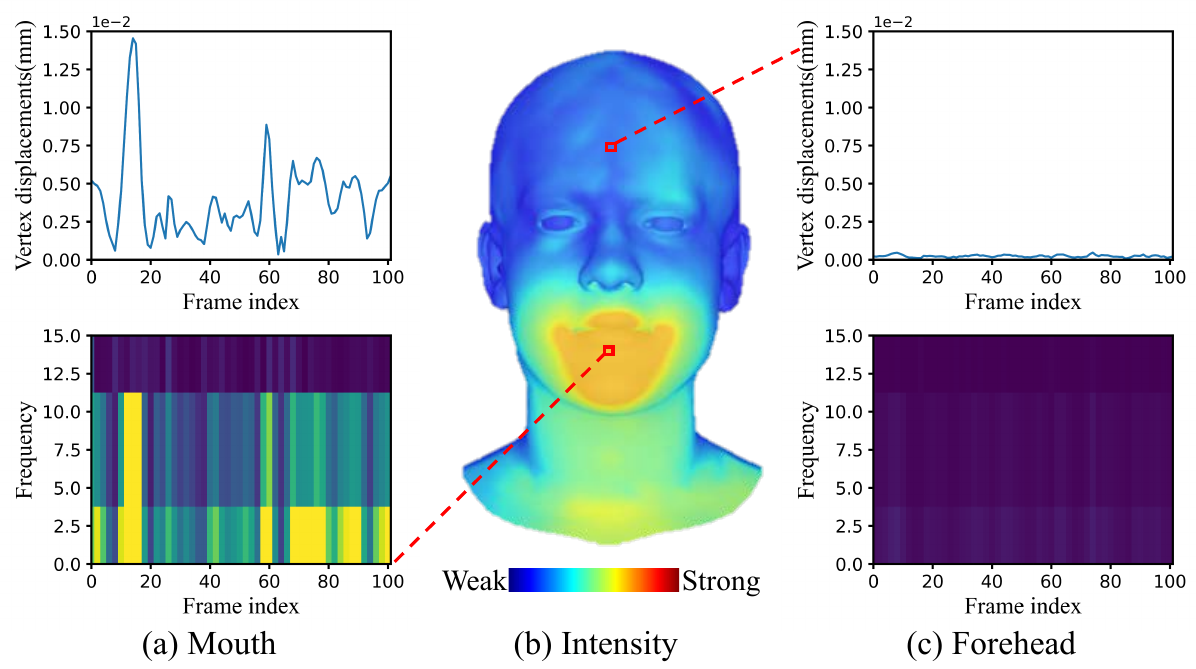}
    \caption{Visualization of differences in facial activity intensity across distinct facial regions on the VOCASET. Vertex activity in the mouth and forehead region within a motion sequence is shown in (a) and (c) (top row: \(L_{2}\) distance between vertices in the reference sequence and the neutral geometry; bottom row:  short-time Fourier transform (STFT) of vertex displacements.). (b) represents the amplitude value of all vertices in the fundamental frequency obtained by the STFT, and facial intensity dynamics in a sequence in the fundamental frequency can be seen in the \textbf{\textit{supplementary video}}. }
    \label{frequency}

\end{figure}

Speech-driven 3D facial animation is a complex and challenging cross-modal task from speech to vision. Some studies investigate speech modality. Earlier works\cite{siatras2008visual,taylor2012dynamic,6585824,weise2011realtime} established a mapping between phonemes and facial gestures. A single phoneme may correspond to several plausible lip shapes, leading to cross-modal uncertainty. For example, in the words 'cafeteria', 'story', and 'expect', the phoneme 'T' is spoken with different lip shapes\cite{taylor2012dynamic}. To alleviate  facial gesture ambiguity, a short sliding window mechanism was introduced to clip several consecutive speech frames\footnote{The speech frame indicates the length of the audio corresponding to a visual frame.}, and then animates the respective visual frame \cite{chai2022speech,cudeiro2019capture,liu2021geometry}. The short audio windows capture additional information from neighboring speech frames but still lead to uncertainty in variations in facial movements. MeshTalk\cite{richard2021meshtalk} applied the long-term audio windows to synthesize each visual frame. FaceFormer\cite{fan2022faceformer} proposed a transformer-based model to capture the dependency of long-term audio context at the frame level. Although capturing long-term context can improve the realistic performance of speech-driven facial animation\cite{karras2017audio}, excessively long-term context inevitably introduces redundant information~\cite{xuefeng}, and long or short single-level speech features lack sufficient temporal resolution. In contrast, some works\cite{taylor2012dynamic,siatras2008visual,10143326,prajwal2020lip} focus only on the animation of the mouth. Mouth movements are most common in speaking activities but other facial gestures cannot be ignored in that speaking activities stimulate synergistic movements of the mouth and other facial muscles\cite{jun2016real,jin2023facial,chen2021talking,vougioukas2019end}. Recently, the frontier has been promoted by directly driving the entire facial animation using a single-level speech features\cite{6585824,weise2011realtime,fan2022faceformer,richard2021meshtalk,cudeiro2019capture,karras2017audio,chai2022speech,ye2023geneface,xing2023codetalker}. However, the differences in facial activity across distinct regions\cite{jun2016real,xia2021local}, such as the mouth and others, were ignored.

\begin{figure}
    \setlength{\textfloatsep}{0pt}
    \centering
    \includegraphics[width=3.45 in]{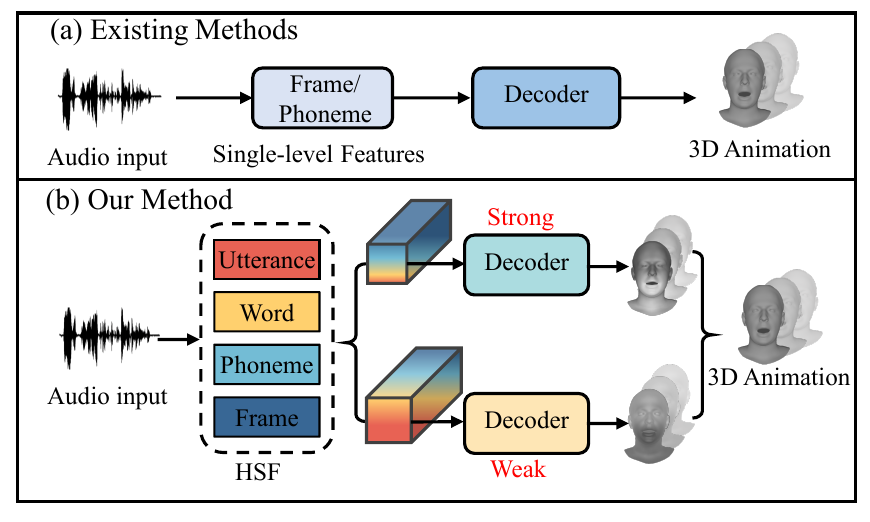}
    \caption{Compared to existing methods (a), our proposed method (b) can accurately generate highly realistic facial animation frames by considering the correlation between HSF and facial activities of varying intensity.}
    \label{fig:figure1}
\end{figure}

The facial activity intensity (FAI) metric is defined to distinguish differences in facial activity, obtained by analyzing the amplitude value within the fundamental frequency of the facial vertex displacements. The FAI across different regions is shown in Fig.~\ref{frequency}. From Fig.~\ref{frequency} (a) and (c), the vertex in the mouth region shows obvious vertex displacements which is the \(L_{2}\) distance between vertices in the reference sequence and the neutral geometry; in contrast, the vertex in the forehead is extremely weak. After short-time Fourier transform (STFT) of the vertex displacements, it is clear that the mouth shows larger amplitude values in the fundamental frequency band than the vertex in the forehead region. Moreover, the movement of the mouth vertex is still strong in the higher frequency bands. Fig.~\ref{frequency} (b) illustrates the amplitude value for the entire face at the fundamental frequency. The mouth displays strong motions, while other facial regions exhibit comparatively weak activity levels. Based on the differences in FAI, we split the activity intensity into two distinct levels: \textbf{Strong} and \textbf{Weak}. To effectively address varying levels of FAI, we introduce a dual-branch decoder to simultaneously synthesize facial animation with varying intensities and ensure a wider intensity facial animation synthesis. Disregarding the differences in FAI could lead to unnatural and oversmoothing facial animation. 

To accommodate differences in FAI, we introduce hierarchical speech features (HSF). According to the inherent characteristics of speech, a hierarchical speech processing architecture divides audio from low to high into the frame-, phoneme-, word-, and utterance-level features\cite{chen2022speechformer,ren2022prosospeech,9383629,liu2020mockingjay}. Compared to single-level speech features, the hierarchical architecture allows for multilevel abstractions of speech, providing a more complementary and comprehensive speech representation with sufficient temporal resolution. HSF has been widely applied in various tasks such as speech emotion recognition\cite{chen2022speechformer,zhao2019automatic,shen2020wise}, automatic speech recognition\cite{ren2022prosospeech} and speech synthesis\cite{lin2020unified,9383629}. However, the hierarchical speech architecture is disregarded in speech-driven 3D facial animation tasks. The facial motions in different regions are related to HSF, where each level of speech representation contributes to animating different facial movements\cite{4067013}. For instance, frame-level features are closely linked to the movements of the lips and jaw, while phoneme-, word- and utterance-level features are associated with broader facial expressions\cite{yu2021multimodal,liu2020mockingjay}. Therefore, the correlation between HSF and facial motions of different intensities needs to be explored.

Based on the abovementioned differences in FAI and the limitations of single-level speech features, we introduce a novel framework in Fig.~\ref{fig:figure1}(b),  which effectively establishes the temporal correlation between HSF and facial activities of different intensities to generate realistic 3D facial animation. Compared to existing methods in Fig.~\ref{fig:figure1} (a), we propose a dual-branch decoding framework to synchronously drive strong and weak facial animation. Additionally, we employ a weighted hierarchical feature encoder to capture the temporal correlation between HSF and facial motions of different intensities. Extensive experiments and a user study indicate that our CorrTalk outperforms existing state-of-the-art methods. The main contributions of this work are listed as follows: 

\begin{itemize}
    \item FAI metric is defined to effectively distinguish different levels of facial activity. Based on the variations in FAI, a dual-branch decoding framework is proposed to synchronously drive strong and weak facial activity.
\end{itemize}
    
\begin{itemize}
    \item We establish temporal correlation between HSF and facial motions of different intensities via a weighted hierarchical feature encoder. 
\end{itemize}

\begin{itemize}
    \item Extensive experiments are performed on two publicly available datasets and demonstrate that our proposed method surpasses existing state-of-the-art approaches. 
\end{itemize}

The remainder of this paper is organized as follows: In Section II, related works, including studies on speech-driven 3D facial animation and the architecture of hierarchical speech processing, are discussed. The design details of the CorrTalk are given in Section III. In Section IV, we present the results of our approach. In Section V, this work is concluded.

\section{Related Work}
In this section, we briefly review previous works on speech-driven 3D facial animation and the architecture of hierarchical speech processing.
\subsection{Speech-Driven 3D Facial Animation}
The automatic synchronization of facial animation with speech is a complex and challenging task that has garnered significant interest in the past few decades. Previous studies focused on generating shapes and movements of the lips or entire facial regions given arbitrary speech conditions. Linguistic-based methods  \cite{aleksic2004speech,6585824,edwards2016jali,taylor2012dynamic,massaro2012animated,xu2013practical,weise2011realtime,zhou2012image,siatras2008visual} bridge a connection between speech and vision via a complex set of mapping rules. For example, the dominance function \cite{massaro2012animated} was proposed to determine facial animation control parameters. Xu \textit{et al.} \cite{xu2013practical} constructed animation curves on a fixed set of static facial poses to synchronize lip and mouth movements. Dynamic visemes \cite{taylor2012dynamic} were introduced to control different lip poses triggered by the same phoneme. JALI \cite{edwards2016jali} considered two anatomical actions, jaw actions and lip actions, to control the 3D facial rig. The above methods allow explicit control of mouth movement accuracy but require much professional manual work and focus only on the movement of the lower face. Data-driven approaches \cite{cao2005expressive,taylor2017deep,zhou2018visemenet} were proposed to synthesize entire 3D facial animation. An animated graph structure and a search algorithm were proposed to predict 3D facial animation \cite{cao2005expressive}. Karras \textit{et al.} \cite{taylor2017deep} developed an end-to-end machine learning model to generate continuous facial animation by using phoneme sequences. VisemeNet, which was proposed by Zhou \textit{et al.} \cite{zhou2018visemenet} utilized a three-stage long short-term memory network to directly predict viseme curves from speech signals. Recently, emotions, expressions, head poses and styles of the speaking head have received considerable attention\cite{zhang20213d,zhang2023sadtalker,xu2023high,ma2023styletalk}. Zhang \textit{et al.} \cite{zhang20213d} utilized audio to produce 3D talking faces with personalized pose sequences. SadTalker\cite{zhang2023sadtalker} predicts 3D motion coefficients of the 3DMM from speech to synthesize accurate facial expressions and head motion in different styles. Xu \textit{et al.} \cite{xu2023high} proposed multimodal emotion space learning to generate arbitrary emotion modalities and generalize to unseen emotion styles. StyleTalk\cite{ma2023styletalk} drives the one-shot portrait to speak with the reference speaking style that is extracted from an arbitrary reference video given another piece of audio.

The most relevant approaches to this work are VOCA \cite{cudeiro2019capture}, MeshTalk \cite{richard2021meshtalk}, FaceFormer \cite{fan2022faceformer} and CodeTalker\cite{xing2023codetalker}, which rely on high-resolution 3D data to directly synthesize face animation in 3D vertex space. VOCA \cite{cudeiro2019capture} extracts audio features by the DeepSpeech model and then generates facial animation with short audio sliding windows. MeshTalk \cite{richard2021meshtalk} considers longer audio contexts and trains a categorical latent space to disentangle audio-correlated and audio-uncorrelated information. FaceFormer \cite{fan2022faceformer} captures the long-term speech context based on a transformer and then employs an autoregressive manner to synthesize facial animation. CodeTalker\cite{xing2023codetalker} introduced a discrete motion priors codebook and generated speech-conditional facial animations based on a code query strategy. However, these works directly drive the entire facial animation using only a single level of speech features while disregarding the differences in FAI across distinct regions. As a result, the generated facial animations may appear unrealistic with overly smoothed movements and a lack of subtle details.

\subsection{Hierarchical Speech Processing Architecture}
Speech signals convey a vast amount of crucial information in everyday conversations. However, adequately representing this information in a digital system is often challenging. Numerous studies \cite{chen2022speechformer,tan2020fine,zhao2019automatic,ren2022prosospeech,lin2020unified,anoop2019automatic} have extracted elusive information from speech based on its natural characteristics. According to inherent speech characteristics, speech signals are decomposed from low to high levels into four levels \cite{chen2022speechformer}, \textit{i.e.,} frame-, phoneme-, word-, and utterance-level. A hierarchical speech processing architecture adopts a multilevel processing approach for speech signals to capture both low-level features and high-level features, thereby improving the performance and robustness of various speech-related tasks. Building on the hierarchical architecture, several works \cite{chen2022speechformer,tan2020fine,zhao2019automatic,ren2022prosospeech,liu2020mockingjay,shen2020wise,lin2020unified} have obtained impressive performances. Tan \textit{et al.} \cite{tan2020fine} employed the phoneme level to predict fine-grained speaking styles. A hierarchical attention transfer network was designed to produce frame level and sentence level features for depression severity measurement \cite{zhao2019automatic}. A hierarchical audio representation is designed at the word-, phoneme-, and frame-levels, which leads to the forming word-level acoustic features that are more relevant to emotion recognition \cite{shen2020wise}. Lin \textit{et al.} \cite{lin2020unified} proposed a single framework to recognize multilingual speech based on phoneme- and word-level features. SpeechFormer \cite{chen2022speechformer} captured a hierarchical speech representation at the frame-, phoneme-, word-, and utterance-level for speech emotion recognition. ProsoSpeech\cite{ren2022prosospeech} introduces a phoneme encoder and word encoder to synthesize speech from text. A hierarchical framework for rhyme modeling, which combines the advantages of phoneme-level and word-level prosody to synthesize speech signals, is proposed\cite{9383629}. 

HSF effectiveness has been widely demonstrated in speech emotion recognition \cite{chen2022speechformer,zhao2019automatic,shen2020wise}, automatic speech recognition \cite{ren2022prosospeech} and speech synthesis \cite{lin2020unified,9383629}. However, in the cross-modal task of speech-driven 3D facial animation, the superiority of HSF is often overlooked. Unlike speech emotion recognition, automatic speech recognition and speech synthesis tasks, speech-driven 3D facial animation is a complex cross-modal task that requires not only capturing the phonetic and prosodic characteristics of speech but also understanding the meaning behind the words being spoken. HSF provides a more comprehensive and complementary representation of speech signals, enabling the animation system to better capture the underlying semantic information and express it through facial movements. Therefore, incorporating HSF is necessary to improve the performance and realism of speech-driven facial animation. Further research in this area can lead to the development of more advanced and effective techniques for speech-driven facial animation.

\begin{figure*}
    \centering
    \includegraphics[width=7.1 in]{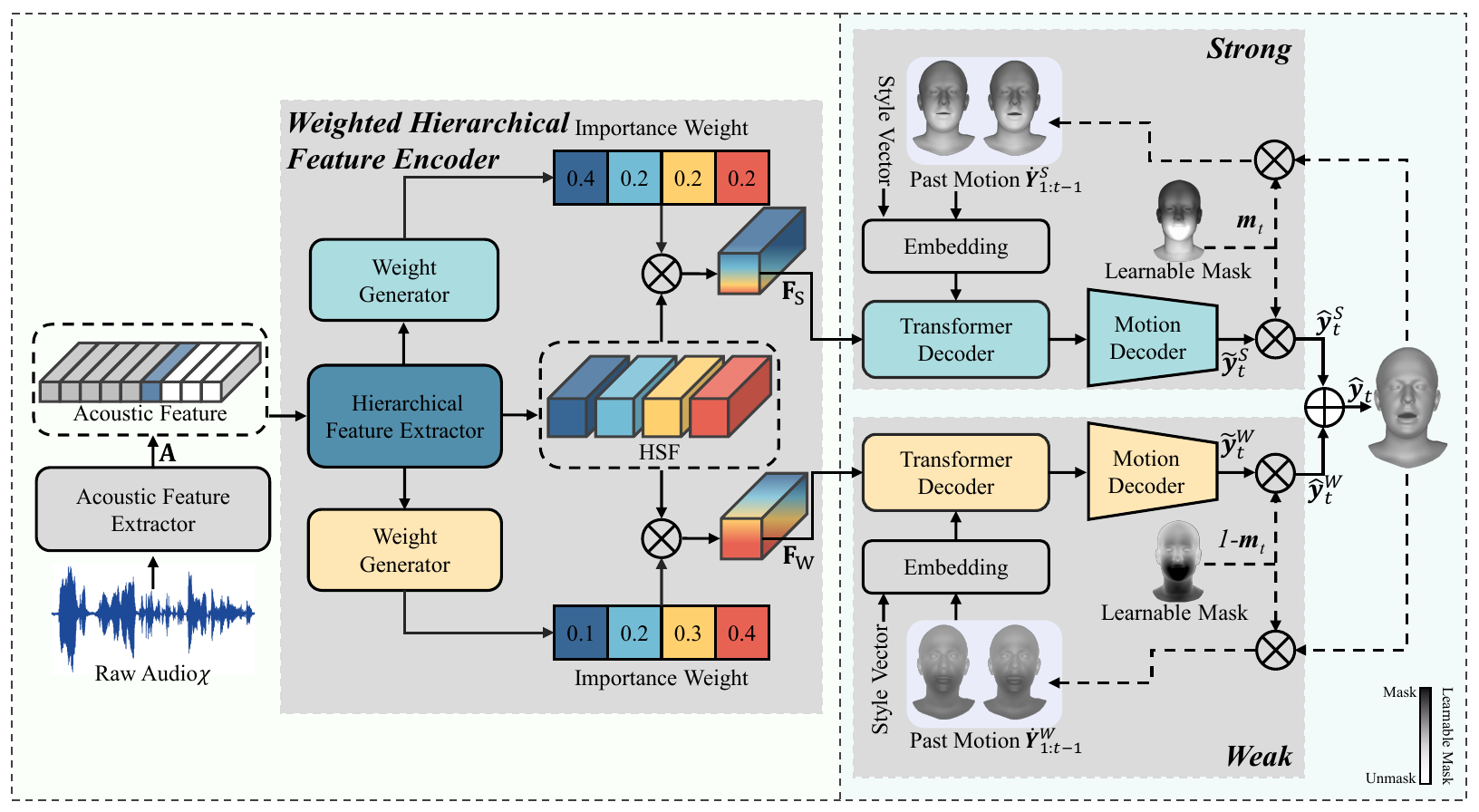}
    \caption{Overview of the proposed CorrTalk. A novel framework for learning the temporal correlation between HSF and facial activities of different intensities uses raw audio as input and generates a sequence of 3D facial animation. The design of the acoustic feature extractor follows wavLM. The weighted hierarchical speech encoder produces frame-, phoneme-, word- and utterance-level speech features, and calculates the importance weight of each level of features for strong and weak facial movements. A dual-branch decoder based on the FAI synchronously generates strong and weak facial movements. After performing STFT of the vertex displacements from training data, a learnable mask \(\mathbf{m}_t \in [0, 1] \) is initialized according to the absolute value of the amplitude in fundamental frequency. \(\mathbf{m}_t (\cdot) \) close to 1 indicates strong facial movements and vice versa for weak movements.}
    \label{framework}
\end{figure*}

\section{Methodology}
In this section, we describe our proposed CorrTalk in detail. First, we briefly overview our proposed framework. Second, we introduce a definition of FAI metric. Then, we describe the key components of CorrTalk: weighted hierarchical feature encoder and dual-branch decoder based on the FAI. Finally, the loss function is illustrated.

\subsection{Overview}

CorrTalk animates arbitrary neutral face geometry under speech signal conditions to generate realistic 3D facial animation. During speaking activity, the mouth region shows strong facial motions, whereas activity in other regions is relatively weak (FAI metric is defined in section ~\ref{FAIR}.). However, differences in FAI across distinct regions are often ignored, leading to smooth facial movements. To deal with these differences, we propose a novel dual-branch decoding framework CorrTalk, in Fig.~\ref{framework}, which effectively establishes the temporal correlation between HSF and facial activities of different intensities to generate realistic 3D facial animation. Specifically, we split facial activity into two categories, strong and weak, and utilize a dual-branch decoder to synchronously synthesize facial activities of different intensities. To accommodate separate decoding of each branch, we propose a weighted hierarchical speech feature encoder to establish temporal correlation between HSF and facial motions of different intensities. 

The CorrTalk framework can be formulated as follows:
a raw audio sequence $\boldsymbol{\chi }$ corresponds to ground truth facial motion sequence $\mathbf{Y} =[\mathbf{y}_{1},\cdots,\mathbf{y}_{t},\cdots, \mathbf{y}_{T} ]\in \mathbb{R}^{T \times V\times 3}$, where $T$ is the number of frames of the facial motion sequence, and $V$ is the number of vertices for each geometry. The raw audio $\boldsymbol{\chi }$ is processed by an acoustic feature extractor, wavLM \cite{chen2022wavlm}, to obtain acoustic features \(\mathbf{A} \in \mathbb{R}^{T \times d_{0}} \), where \(d_0\) denotes the feature dimension of the acoustic feature extractor output. The output frequency of the raw wavLM is a fixed value that fails to be directly edited, which could result speech feature frequency being different from facial animation frequency. Therefore, the linear interpolation layer is embedded in wavLM to adjust the frequency of the acoustic features \cite{fan2022faceformer}, namely, \( \mathrm{f}_{a}=\mathrm{n}\cdot \mathrm{f}_{m}, \mathrm{n} \in \mathbb{N}^+ \), where \( \mathrm{f}_{m}\) is the frequency of facial sequences and \( \mathrm{f}_{a}\) represents acoustic feature frequency. Then, a weighted hierarchical feature encoder is employed to obtain weighted hierarchical features \( \mathbf{F}=[\mathbf{F}_{S}, \mathbf{F}_{W}]\). The equation that represents this process is:

\begin{equation}
\begin{gathered}
\mathbf{A} = \rm Acoustic\; Feature\; Extractor_{\theta}(\boldsymbol{\chi });\\
\mathbf{F} = \rm Weighted\; Hierarchical\; Feature\;Encoder_{\phi}(\mathbf{A}),
\end{gathered}
\end{equation}
\noindent where \( \theta \) and \( \phi \) represent the model parameters of the acoustic feature extractor and weighted hierarchical feature encoder, respectively, and \(\mathbf{F}_{S} \in \mathbb{R}^{T \times d}\) and \(\mathbf{F}_{W} \in \mathbb{R}^{T \times d} \) are used for cross-modal decoding of strong and weak branches, respectively, where \(d\) represent the speech feature dimension. The final goal is to generate a facial motion sequence $\mathbf{\hat{Y} } =[\mathbf{\hat{y}} _{1},\cdots,\mathbf{\hat{y}} _{t},\cdots, \mathbf{\hat{y}} _{T} ]\in \mathbb{R}^{ T \times  V\times 3} $, which is similar to $\mathbf{Y} $. The formula is expressed as:

\begin{equation}
\begin{gathered}
\mathbf{\hat{Y}}^S = \rm Strong\; Branch_{\varphi}(\mathbf{F}_{S});\\
\mathbf{\hat{Y}}^W = \rm Weak\; Branch_{\psi}(\mathbf{F}_{W});\\
\mathbf{\hat{Y}} = \mathbf{\hat{Y}}^S + \mathbf{\hat{Y}}^W ,\\
\end{gathered}
\end{equation}
\noindent where \(\mathbf{\hat{Y}}^S \) and \(\mathbf{\hat{Y}}^W\) are the outputs of the strong and weak branches,respectively, and \( \varphi \) and \( \psi \) represent the model parameters of the strong and weak cross-modal decoding branches, respectively. In the following, more detailed descriptions of each component are provided.

\subsection{FAI Metric} 
\label{FAIR}
To distinguish the differences in facial activity, we define the FAI metric. The sequence of vertex displacements \( \mathbf{D}_{ver} \in  \mathbb{R}^{ T \times V} \) is a key component in the analysis of dynamic facial activities, generated by computing the \( {L}_{2}\) distance between facial vertices in the reference sequence  $\mathbf{Y} \in \mathbb{R}^{T \times V\times 3} $ and the neutral face geometry \( \mathbf{h} \in \mathbb{R}^{ V  \times 3}\). This process records the spatiotemporal deviations of facial vertices from their neutral geometry, effectively representing the dynamic evolution of facial motions. To unveil the changing frequency and amplitude characteristics of facial motions, we employ STFT to decompose \( \mathbf{D}_{ver}\) into its constituent frequency components over time. The process is expressed as:
\begin{equation}
\begin{aligned}
\label{intensity}
\mathbf{D}_{ver}= \left \| \mathbf{Y} -\ \mathbf{h} \right \|_{2}^{2}; \\
\mathbf{I} = STFT(\mathbf{D}_{ver}),
\end{aligned}
\end{equation}
\noindent where \( \mathbf{I} \in \mathbb{R}^{ b \times T \times V} \) represents the changes in amplitude across \(b\) frequency bands. 

In this work, we utilize the amplitude value of the fundamental frequency as a metric for quantifying the FAI. Specifically, a larger amplitude value corresponds to strong facial activity and vice versa for weak facial activity. Combined with Fig.~\ref{frequency} , it becomes evident that regions encompassing the lips and mouth exhibit strong facial motions. In contrast, areas such as the forehead and cheeks display comparatively weak facial motions.

\subsection{Weighted Hierarchical Feature Encoder}

The weighted hierarchical features encoder provides hierarchical speech features to synchronously adapt to strong and weak facial activity, including a hierarchical feature extractor, and two weight generators.

\begin{figure}
    \setlength{\textfloatsep}{0pt}
    \centering
    \includegraphics[width=3.5 in]{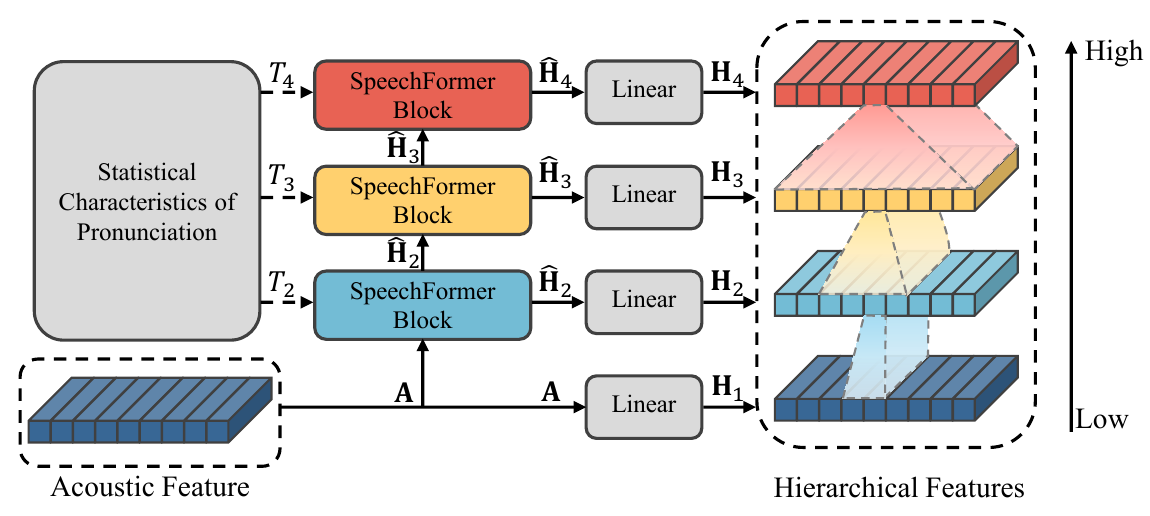}
    \caption{Overview of hierarchical feature extractor. }
    \label{fig:HSFE}
\end{figure}

\noindent \textbf{Hierarchical Feature Extractor.} 
Utilizing the acoustic features \(\mathbf{A}\) as the input, the hierarchical feature extractor in Fig.~\ref{fig:HSFE} provides more comprehensive speech representation across various levels of granularity, including frame-, phoneme-, word- and utterance-level. Specifically, the frame-level speech features \( \mathbf{H}_{1} \in \mathbb{R}^{T \times d }\) are derived by projecting \( \mathbf{A}\) via a linear layer. Additionally, the SpeechFormer block is employed to extract phoneme-grained features \(\mathbf{\hat{H}}_2 \in \mathbb{R}^{T \times d_{0} }\) based on both phoneme duration and \( \mathbf{H}_{1} \). Subsequently, \(\mathbf{\hat{H}}_2\) undergoes linear projection, resulting in the acquisition of phoneme-level features \( \mathbf{H}_{2} \in \mathbb{R}^{T \times d }\). Similarly, the process extends to generate word-level features \( \mathbf{H}_3 \in \mathbb{R}^{T \times d } \) and utterance-level features \( \mathbf{H}_4 \in \mathbb{R}^{T \times d } \). The core of the hierarchical feature extractor is the SpeechFormer block, which is based on statistical characteristics of the pronunciation structure to capture the correlation between adjacent tokens. Compared to the classic transformer, the SpeechFormer block does not calculate the multihead attention of all tokens but captures the multihead attention of tokens in the statistical duration of a pronunciation structure. \( T_{2} \), \( T_{3} \) and \( T_{4} \) indicate the statistical duration of the phoneme, word, and utterance, respectively.  The length of the phoneme is variable from 50 ms to 200 ms, and the duration of the word is from 250 ms to 1000 ms. Note that utterance duration is set to its input sequence time.  More details about the SpeechFormer block are described in \cite{chen2022speechformer}. To ensure that the HSF at each level has the same number of tokens, an overlapping sliding-window strategy is employed with a stride of one in this work. 

Existing approaches have primarily relied on frame-level speech features \( \mathbf{H}_{1} \) to directly drive facial movements, resulting in the omission of considerable information from other linguistic units, such as phonemes, words, and utterances. Although some studies have used sliding windows or attention with the transformer mechanism to obtain correlation between adjacent frames at a single level, these speech features suffer from insufficient temporal resolution, limited expressiveness, and a lack of context. Based on the statistical characteristics of the pronunciation structure, the hierarchical feature extractor provides more comprehensive and complementary speech representations with sufficient temporal resolution.

\noindent \textbf{Weight Generator.} HSF can provide more comprehensive and complementary speech representations. Specifically, the low-level features supply more detail to complement the high-level speech features, and the high-level features give semantic contextual information to the low-level features. To encode the temporal correlation between HSF and facial animations of different intensities, the weight generator is proposed. For any \( k \in [S, W] \) cross-modal decoding branch, the importance weight \( \hat{\bm{\alpha}}_{k} \in \mathbb{R}^{T \times 1 } \) of frame-level features \( \mathbf{H}_1 \) is described as:

\begin{equation}
\begin{gathered}
\mathbf{Q}_1 = \Phi (\mathbf{H}_1); 
\mathbf{K}_1 = \Psi (\mathbf{H}_1);  \\
\mathbf{Z}_1 = tanh(\mathbf{Q}_1+\mathbf{K}_1); \\
\hat{\bm{\alpha}}_{k} =  \Theta(\mathbf{Z}_1), 
\end{gathered}
\end{equation}

\noindent where \( \Phi\), \( \Psi\) and \( \Theta\) are linear projection operation, \( \{ \mathbf{Q}_1, \mathbf{K}_1, \mathbf{Z}_1 \} \in \mathbb{R}^{T \times d_{1} } \). Similarly, the importance weights of phoneme-, word-, and utterance-level features \( (  \hat{\bm{\beta}}_{k}, \hat{\bm{\gamma}}_{k} \)  and  \( \hat{\bm{\delta}}_{k} ) \) are obtained. A softmax function is introduced to assign the final importance weight of the hierarchical features.

\begin{equation}
[ \bm{\alpha} _{k}, \bm{\beta} _{k}, \bm{\gamma} _{k}, \bm{\delta}_{k}] = softmax([\hat{\bm{\alpha}}_{k}, \hat{\bm{\beta}}_{k}, \hat{\bm{\gamma}}_{k}, \hat{\bm{\delta}}_{k}]).
\label{equ2}
\end{equation} 

\noindent Based on the weights of each level features, HSF is integrated to obtain weighted hierarchical features \(\mathbf{F}_{k}\) for decoding facial animations.
\begin{equation}
\mathbf{F}_{k} = \bm{\alpha} _{k}  \mathbf{H}_1 + \bm{\beta} _{k}  \mathbf{H}_2 +  \bm{\gamma} _{k}  \mathbf{H}_3 + \bm{\delta}_{k} \mathbf{H}_4.
\label{equ3}
\end{equation} 

\noindent The weighted hierarchical features \( \mathbf{F}_S\) and \( \mathbf{F}_W\) are finally obtained for cross-modal decoding.

\subsection{Dual-branch Decoder Based on FAI}
\label{section:cross-frequency}
Due to the differences in FAI, it is difficult for a single decoder to accommodate the entire facial movements. Thus, a dual-branch cross-modal decoder based on the intensity of facial activity is developed, mapping from  \( \mathbf{F}_S\) and \( \mathbf{F}_W\) to the strong and weak facial motions, respectively. 

As shown in the right part of Fig.~\ref{framework}, each decoding branch consists of an embedding layer, a transformer decoder and a motion decoder with a linear layer. For the strong motion decoding branch,  the output \( \mathbf{\tilde{y}}_{t}^{S} \) of the motion decoder is obtained based on a temporal autoregressive way under the condition of the talking style vector, the weighted hierarchical features $\mathbf{F_S}$ and past motions \( \mathbf{\dot{Y}}_{1:t-1}^{S} \).  \( \mathbf{\dot{Y}}_{1:t-1}^{S} \) can be written as:

\begin{equation}
\begin{gathered}
\mathbf{\dot{Y}}_{1:t-1}^{S} = \mathbf{m}_{t} \mathbf {\Hat{Y}}_{1:t-1};\\
\mathbf{m}_{0} = \frac{1}{T} \sum_{t=1}^{T} \left\| \mathbf{I}_{0(tj)} \right\|  ,j=(1,\cdots,V),
\end{gathered}
\label{equ5}
\end{equation}

\noindent where \(\mathbf{m}_{t} \in [0, 1] \) is a learnable mask and initialised with a normalized \(\mathbf{m}_0 \in \mathbb{R}^{1 \times V}\), and \(\mathbf{I}_0 \in \mathbb{R}^{T \times V} \) represents the amplitude value of FAI metric \(\mathbf{I}\) (in Eq.~\ref{intensity}) in the fundamental frequency band. Values in \(\mathbf{m}_{t}\) close to 1 indicate strong facial movements and vice versa for weak movements. Similarly, for the weak motion decoding branch, \(\mathbf{\tilde{y}}_{t}^{W}\) is obtained in the same way. However, there are some differences, and this branch focuses on weak facial motions. Thus, we replace \(\mathbf{m}_{t}\) in Eq.~\ref{equ5} with \(1-\mathbf{m}_{t}\) to ensure intra-frame consistency.

The final predicted motion \( \mathbf{\hat{y}}_{t}\) can be described as:

\begin{equation}
\mathbf{\hat{y}}_{t}=\mathbf{m}_{t}  \mathbf{\tilde{y}}_{t}^{S}  + (1-\mathbf{m}_{t})  \mathbf{\tilde{y}}_{t}^{W} .
\end{equation} 

\noindent The newly predicted motion \( \mathbf{\hat{y}}_{t} \) is employed to update the past motions as \(\mathbf {\Hat{Y}}_{1:t-1}\) to prepare for the next motion. The above decoding process is performed recursively until the complete sequence animation is predicted.

\subsection{Loss Function}
To develop CorrTalk, the loss function includes a reconstruction loss \(\mathcal{L}_{rec}\), and a velocity loss \(\mathcal{L}_{vel}\). The final loss function \( \mathcal{L}_{total}\) is presented as:
\begin{equation}
\mathcal{L}_{total}=\mathcal{L}_{rec}+ \mathcal{L}_{vel}.
\end{equation}

\noindent \textbf{Reconstruction Loss}:  \( \mathcal{L}_{rec} \) is represented as:
\begin{equation}
\mathcal{L}_{rec}=\sum_{t=1}^{T}\sum_{v=1}^{V} \left \| y_{t,v}-\hat{y}_{t,v} \right \|_{2}^{2},
\end{equation} 
\noindent where \(y_{t,v}\) and \( \hat{y}_{t,v} \) are the ground truth and synthetic geometry, respectively, at the \( v^{th}\) vertex in the \( t^{th} \)frame and \( \left \| \cdot  \right \| _{2}^{2} \) indicates the \( L_{2}\) distance.

\noindent \textbf{Velocity Loss}:  Although an autoregressive strategy is adopted to ensure temporal continuity, humans are sensitive to high-frequency jitter, and other subtle disturbances may interfere with strong motion synthesis. Thus, to enhance the realism of the facial motions, the velocity loss \( \mathcal{L}_{vel}\) is proposed and defined as:
\begin{equation}
\begin{gathered}
\mathcal{L}_{vel}=\sum_{t=2}^{T} \left \| \mathbf{m}_{t} (\textbf{y}_{t}-\textbf{y}_{t-1})- \mathbf{m}_{t} (\mathbf{\hat{y}}_{t} -\mathbf{\hat{y}}_{t-1} ) \right \|_{2}^{2}.
\end{gathered}
\end{equation}

\section{Experiments and Results}
In this section, we report the experiments performed on two available corpora. First, we introduce the details of the experimental data and implementations. Second, we compare our method with previous works and evaluate the experimental results by quantitative evaluation, qualitative evaluation, and a user study. Finally, we conduct extensive ablation studies to evaluate the effectiveness of key components.

\subsection{Dataset and Implementations}
In this work, two publicly available datasets, BIWI \cite{fanelli20103biwi} and VOCASET \cite{cudeiro2019capture},  were employed to train and evaluate different methods. Both datasets provide audio-3D head geometry pairs of spoken English utterances. BIWI contains 40 unique sentences shared across all speakers in the dataset. VOCASET contains 255 unique sentences, some of which are shared among different speakers.

\noindent \textbf{BIWI Dataset.} BIWI is an audio-visual corpus with corresponding dense dynamic 3D face geometries. Fourteen human subjects, eight females and six males, were asked to orally articulate 40 English sentences, each of which was then recorded on two occasions, once in a neutral context and once in an emotional context. These 3D face geometry sequences were captured at 25 fps, each geometry with 23370 vertices. We follow FaceFormer\cite{fan2022faceformer} and CodeTalker \cite{xing2023codetalker} in using only the subset with emotions.
Specifically, the dataset is divided into a training set (BIWI-Train), a validation set (BIWI-Val), and two test sets (BIWI-Test-A and BIWI-Test-B). BIWI-Train, BIWI-Val, BIWI-Test-A, and BIWI-Test-B include 192 sequences (6 subjects × 32 sentences), 24 sequences (6 subjects × 4 sentences), 24 sequences (6 subjects × 4 sentences), and 32 sequences (8 unseen subjects × 4 sentences), respectively.

\noindent \textbf{VOCASET Dataset.} VOCASET contains 480 audio-3D geometry pairs from 12 subjects. For each subject, 40 sequences of 3 \( \sim  \) 5 seconds in length were captured at 60 fps. A 3D head geometry is represented by 5023 vertices and 9976 faces. In this study, the frequency of 3D head geometry is downsampled to 30 fps by referring to the source code\footnote{Source code:\url{https://github.com/EvelynFan/FaceFormer}} of FaceFormer \cite{fan2022faceformer} for fair comparisons. The dataset is divided into a training set (VOCASET-Train) of 320 sequences (8 subjects × 40 sentences), a validation set (VOCASET-Val) of 40 sentences (2 unseen subjects × 20 sentences) and a test sets (VOCASET-Test) of 40 sentences (2 unseen subjects × 20 sentences).

\noindent \textbf{Implementations.} Four state-of-the-art methods were employed for comparison with our method, including VOCA \cite{cudeiro2019capture}, MeshTalk \cite{richard2021meshtalk}, FaceFormer \cite{fan2022faceformer} and CodeTalker\cite{xing2023codetalker}. Some of these methods do not provide pretrained models on BIWI or VOCASET, and we retrain a model on VOCASET or BIWI following the official source code. For example, MeshTalk needs to retrain on BIWI and VOCASET. For fair comparison, we perform the same configurations as VOCA\cite{cudeiro2019capture}, FaceFormer\cite{fan2022faceformer} and CodeTalker\cite{xing2023codetalker} to divide the datasets BIWI and VOCASET. 

We train CorrTalk on a single NVIDIA A100 for 100 epochs. The model parameters are updated using the Adam optimizer with \(beta_{1} = 0.9\), \(beta_{2} = 0.999\), and the base learning rate = \( 10^{-4} \). The learning rate decays to \(50\% \) of the original rate every 80 epochs. The linear layer conversion projects the 1024-dimensional speech-grained representations to the \( d \) dimension with \( d=128\) on BIWI and \(d=64\) on VOCASET. In the weight generator, \(  d_{1}\) is set to 32.

\subsection{Quantitative Evaluation}
\label{section:Quantitative Evaluation}
Following MeshTalk, FaceFormer and CodeTalker, the lip vertex error \cite{richard2021meshtalk,fan2022faceformer, xing2023codetalker} is employed to measure the quality of lip movements. The maximum \( L_{2} \) error is calculated for all lip region vertices for each frame, and then the average of all frames is reported. Moreover, we introduce upper-face dynamics deviation (FDD) \cite{xing2023codetalker} to evaluate the facial dynamics compared to the ground truth. 

We calculate the quantitative evaluation metrics, lip vertex error and FDD, for all sequences in the BIWI-Test-A, and take the average results for comparison purposes. As shown in Table~\ref{quantitative}, the proposed CorrTalk provides the lowest lip vertex error of 4.0858 and FDD of 2.8359, surpassing the existing state-of-the-art methods. The results obtained by our method are attributed to the following three reasons: (1) The dual-branch decoding strategy based on FAI synchronously synthesizes strong and weak facial animation, avoiding high-frequency details being smoothed and subtle movements being ignored.  (2) HSF provides a more comprehensive representations of speech signals and contextual information with sufficient temporal resolution. (3) By calculating the importance weight of each level of features to facial activities of different intensities, the weighted hierarchical features can effectively accommodate facial activity of different intensities, producing natural and expressive facial animations. In summary, we proposed a method that improves the quality of speech-driven facial animation by leveraging a comprehensive representation of speech, considering the characteristics of FAI and capturing the correlation between HSF and facial activities of different intensities.

\begin{table}[h]
\caption{Quantitative evaluation on BIWI-Test-A. For both metric, lower is better.}
\resizebox{\linewidth}{!}{
\begin{tabular*}{0.8\linewidth}{@{}lcc@{}}
\toprule
\multirow{2}{*}{Methods} & Lip Vertex Error                 & FDD                      \\
                         & (\(\times 10^{-4}\)mm)           &(\(\times 10^{-5}\)mm)  \\ \midrule
VOCA                     & 6.5563                           & 8.1816                   \\
MeshTalk                 & 5.9181                           & 5.1025                   \\
FaceFormer               & 5.3077                           & 4.6408                   \\
CodeTalker               & 4.7914                           & 4.1170                   \\
CorrTalk (Ours)        & \textbf{4.0858}                  & \textbf{2.8359}          \\ 
\bottomrule
\end{tabular*}
}
\label{quantitative}
\end{table}

\begin{figure*}
    \centering
    \includegraphics[width=7.2 in]{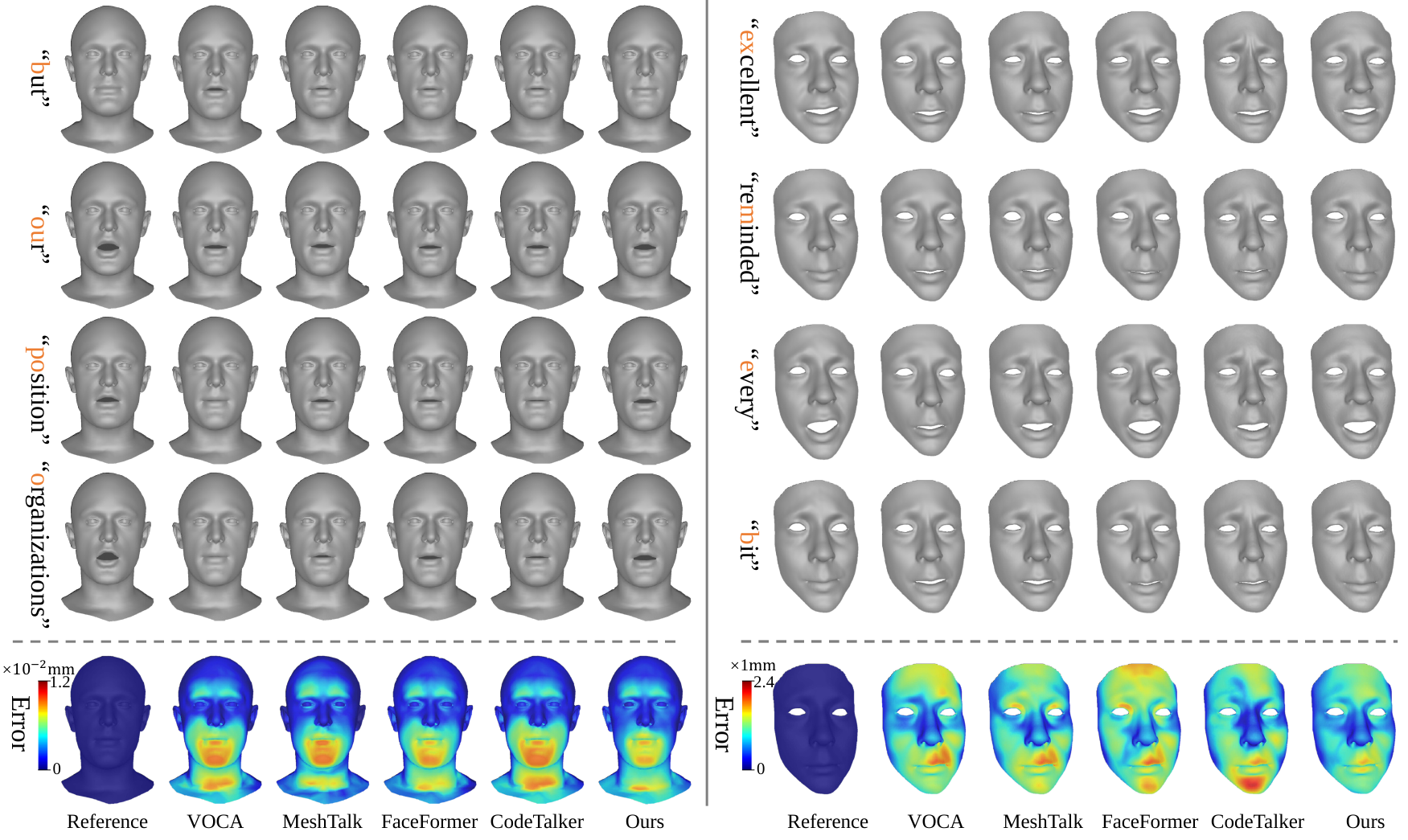}
    \caption{Visual comparison of sampled facial animations generated by different methods on VOCA-Test (left) and BIWI-Test-B (right). The top portion delineates facial animations associated with distinct speech content. The bottom portion displays the synthetic sequence with ground truth mean error.}
    \label{fig_visual_frame}
\end{figure*}

\begin{figure*}[h]
    \centering
    \includegraphics[width=7.2 in]{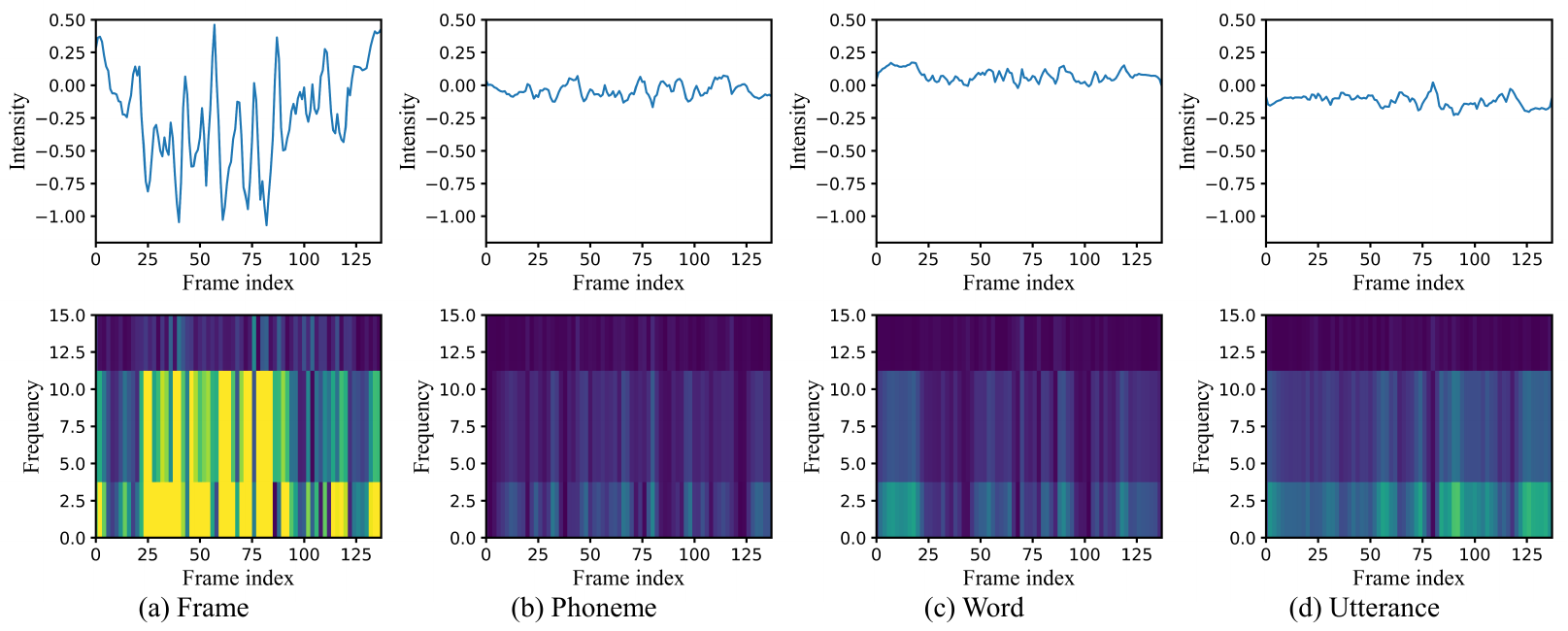}
    \caption{Visual comparison of frame- (a), phoneme- (b), word- (c) and utterance-level (d) speech features. The top row shows the variation in the intensity of speech features in a sequence; the bottom row represents the result of applying the STFT to the feature intensity from the top row.}
    \label{feature_frequency}
\end{figure*}

\subsection{Qualitative Evaluation}

We carry out a qualitative evaluation by visualizing our method with other competitors in Fig.~\ref{fig_visual_frame}. For a fair comparison, we randomly sampled the talking style to VOCA, FaceFormer, CodeTalker and our CorrTalk as the same conditional input. Given the biological nature of the visual system, the human eye is extremely sensitive to motion in a scene, and can quickly and accurately capture motion in the surrounding environment. During conversational activities, the lips have more rapid and significant movements than other regions in Fig.~\ref{frequency}. Thus, synthesizing accurate and realistic lip movements is very challenging, and lip movement is most likely to be noticed by the listener. To evaluate lip synchronization performance, we listed four typical synthetic facial animation frames that speak on specific syllables in the top portion of Fig.~\ref{fig_visual_frame}. The lip movements synthesized by our CorrTalk are more accurate compared to other competitors, and are consistent with the performance of the reference frames. For example,  CorrTalk produces a more accurate lip synthesis with reasonable lip closure when pronouncing bilabial sounds /b/ and /m/ (\textit{i.e.}, in cases "but", "reminded" and "bit"). In addition, our CorrTalk delivers more noticeable and plausible mouth movements when the words "our", "organizations" and "every" are spoken.

Although lip movements are easily perceived, overall facial activity is essential for human communication and expression. For a more comprehensive evaluation, we visualize the average error between the synthesized sequences and the ground truth in the bottom portion of Fig.~\ref{fig_visual_frame}. For all methods, the error in the mouth region is significant compared to other regions. That is, the errors in the synthesized sequences are mainly concentrated in the regions with significant and rapid facial motions. The result suggests that synthesizing accurate mouth movements is challenging. Compared to other competitors' methods, CorrTalk demonstrates minimal overall errors, especially in the mouth and cheeks regions. This is mainly attributed to the fact that the cross-modal decoding strategy based on FAI adequately accounts for the differences in facial movements, and the fact that the HSF provides complementary and sufficiently temporal-resolution representations for facial movements. We encourage readers to watch the animated comparisons in the supplementary video.

\subsection{User Study}

We perform a user study to evaluate the quality of the facial animation generated by our CorrTalk in terms of lip sync and realism. Specifically, we conduct A/B tests on BIWI-Test-B and VOCASET-Test. For BIWI-Test-B, 30 samples are randomly selected for each comparison, i.e., ours vs. competitors. To ensure a fair comparison, these samples evenly cover all talking styles in BIWI-test-B, i.e., each sample is used to synthesize the face animation in the same talking style for VOCA, FaceFormer, CodeTalker and CorrTalk. Thus, we obtained 150 A vs. B pairs (30 samples × 5 comparisons). For VOCASET-Test, we use the same configuration as VOCASET-Test to select 150 evaluation pairs (30 samples × 5 comparisons). In this work, we recruited 30 participants with strong visual and auditory acuity to participate in the user study. Each pair was independently evaluated by at least three different participants, obtaining in a total of 450 entries for BIWI and 450 entries for VOCASET.

The percentages of the assessment results are listed in Table~\ref{user_study}. This outcome corresponds in line with the quantitative evaluation findings in Section\ref{section:Quantitative Evaluation}. The facial animation produced by our proposed method exhibits visual quality superior to existing methods, as evidenced by the generation of subtle movements, precise lip synchronization, and authentic facial expressions. The user study provides empirical evidence for the superior perceptual quality of facial animation. 

\begin{table}[h]
\caption{User study results on BIWI-Test-B and VOCASET-Test.}

\begin{tabular}{ccccc}
\toprule
\multirow{2}{*}{Ours vs. Competitors} & \multicolumn{2}{c}{BIWI-Test-B}  & \multicolumn{2}{c}{VOCASET-Test} \\ \cline{2-5} 
                                      & Lip Sync      & Realism       & Lip Sync      & Realism       \\ \midrule
Ours vs. VOCA                         & 90.20         & 89.22         & 90.32         & 89.25         \\
Ours vs. MeshTalk                     & 85.29         & 80.39         & 93.55         & 91.94        \\
Ours vs. FaceFormer                   & 78.43         & 73.53         & 76.34         & 76.43          \\
Ours vs. CodeTalker                   & 72.55         & 66.67         & 63.33         & 60.00         \\
Ours vs. Ground truth                 & 41.90         & 48.57         & 43.33         & 47.78          \\ \bottomrule
\end{tabular}

\label{user_study}
\end{table}

\subsection{Visualization Analysis for CorrTalk Process}
\label{visual_analysis}
To better understand and analyze CorrTalk, we visualize the HSF, the importance weight of HSF for strong and weak branches, and the output of the dual-branch decoder. 

\subsubsection{HSF} We first list the spectrograms of each of level speech features and perform STFT analysis on the spectrograms in Fig.~\ref{feature_frequency}. In HSF, the frame-level features occupy a wider range of intensities with significant variations. After STFT, it is easier to observe that frame-level features have the strongest speech energy in all frequency bands compared to other level features. Frame-level features show significant speech activity in the high frequency bands. This means that higher frequency and more intense activity can be produced by the frame-level features. In addition, higher-level features in the equal frequency band show longer activity durations as the level of speech features increases. Higher-level speech features focus on the representation of longer-term, higher-level semantic information. HSF provides a more comprehensive and complementary multilevel speech representation, and contextual information with sufficient temporal resolution.

\subsubsection{Importance Weight of HSF}
Although the hierarchical feature extractor produces rich and comprehensive speech representations, the importance weight of each level features to facial movements is difficult to assign. Thus, the weight generator is introduced to capture the correlation between speech representations and facial activity.  Fig.~\ref{atten} visualizes feature importance weight at each level to the facial motions. For the strong motion decoding branch, frame-level features accounted for the highest proportion, followed by phoneme- and word-level features, with utterance level features having the lowest weight. In the weak motion decoding branch, the importance weights of phoneme-, word- and utterance-level features are close and occupy a high share, however frame-level features account for the least share. This phenomenon suggests that there are differences in the weight of speech features at each level to the various facial activities, with some of these features contributing very little, but being essential. Frame-level features endeavor to convey high-frequency details and strong facial activity, whereas  phoneme-, word-, and utterance-level features provide longer-term speech context information, which improves facial animations realism. 

Based on the importance weight produced by the weight generator, the weighted hierarchical features \( \mathbf{F}_{S}\) and \( \mathbf{F}_{W}\) for the strong and weak branches are finally obtained. We visualized  \( \mathbf{F}_{S}\) and \( \mathbf{F}_{W}\) in Fig.~\ref{fusion_feature}. The \( \mathbf{F}_{S}\) is closer to the frame-level features, with more drastic variations and higher-frequency representational intensity. \( \mathbf{F}_{W}\) exhibit weaker speech activity intensity. The hierarchical feature encoder provides heterogeneous speech features for facial activity in each branch. 

\subsubsection{Dual-branch Decoder}
To explain the dual-branch decoding strategy, we visualize the synthetic sequences \( \mathbf{\tilde{Y}}^{S} \), \( \mathbf{\tilde{Y}}^{W} \) and \(\mathbf{\hat{Y}} \) with the ground truth mean error in Fig.~\ref{merge_error}. For the mouth region, the weak decoding branch obtains significant motion errors. In other regions, the output of the weak branch is closer to the ground truth. This is because each branch focuses on different intensities of facial activity. Although there are regions of significant deviation from the reference motions in both branches, the learned mask is effective in both reducing the deviation of the motion and ensuring that the outputs of one branch have similar motion intensity. Our proposed dual-branch decoding strategy effectively balances differences in facial movement intensity. This further confirms the need to consider differences in FAI.

\subsection{Ablation Studies}
In this section, we analyze a series of key components composed of our CorrTalk, including a hierarchical feature encoder, a dual-branch decoding strategy based on FAI, and an acoustic feature extractor. All ablation studies are performed in the same configuration as CorrTalk, except for the component under investigation, and the lip vertex error and FDD are reported on BIWI-Test-A.

\begin{figure}
    \centering
    \includegraphics[width=3.5in]{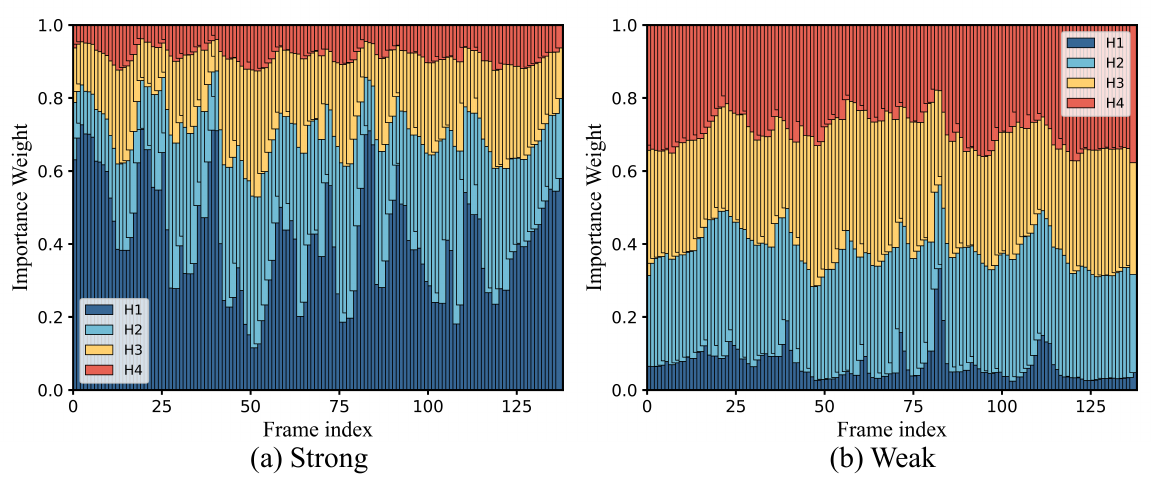}
    \caption{Visualizes the HSF importance weight provided by the weight generator for the strong (a) and weak (b) decoding branches.}
    \label{atten}
\end{figure}

\begin{table}[h]
\centering
\caption{Results of ablation studies for our components on BIWI-Test-A.}
\label{ablation}
\resizebox{\linewidth}{!}{
\begin{tabular*}{\linewidth}{lcc}

\toprule
 \multirow{2}{*}{}  & Lip Vertex Error                & FDD        \\
                    & (\(\times 10^{-4}\)mm)   & (\(\times 10^{-5}\)mm)  \\ 
\midrule
CorrTalk             & 4.0858                    & 2.8359   \\ 
\midrule
\begin{tabular}[c]{@{}l@{}}remove weighted\\ hierarchical features encoder\end{tabular}      & 4.1487                   & 3.7123           \\ 
use single branch                      & 4.5201                   & 3.2039           \\
initialise \(m_{t}\) with  a random way & 4.4987                   & 3.6363           \\ 
replace wavLM with wav2vec 2.0          & 4.5150                   & 3.1968           \\ \bottomrule
\end{tabular*}
}
\end{table}

\subsubsection{Effectiveness of the Weighted Hierarchical Feature Encoder}

To verify the effectiveness of the weighted hierarchical feature encoder, we remove the weighted hierarchical feature encoder and retrain the framework. As seen in Table~\ref{ablation}, the results are unsatisfactory, especially for the FDD increasing from 2.8359 to 3.7123. This phenomenon can be explained by the difficulty of acoustic features alone to provide the comprehensive semantic representations needed for upper facial motions. Combined with Fig.~\ref{atten} (b), frame-level features occupy a lower share; however, phoneme-, word- and utterance-level representations are committed to producing weak facial movements. The frame-level features is typically used to express transient variation in sound and pitch in speech, and they are useful for producing rapid and significant mouth movements. Phoneme-, word- and utterance-level features with longer contextual information, contribute to the overall facial motions. In summary, the weighted hierarchical feature encoder is effective in establishing temporal correlation between HSF and facial motions of different intensities, and is critical for HSF to synchronously accommodate strong and weak facial activities.

\begin{figure}
    \centering
    \includegraphics[width=3.5in]{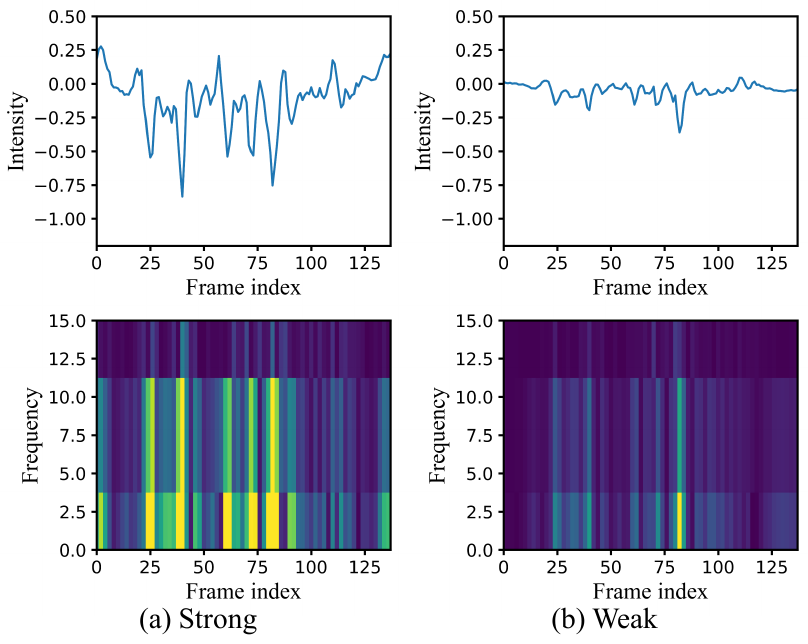}
    \caption{Visualization of weighted hierarchical features in strong (a) and weak (b) decoding branches.}
    \label{fusion_feature}
\end{figure}

\begin{figure}[h]
    \centering
    \includegraphics[width=3.5in]{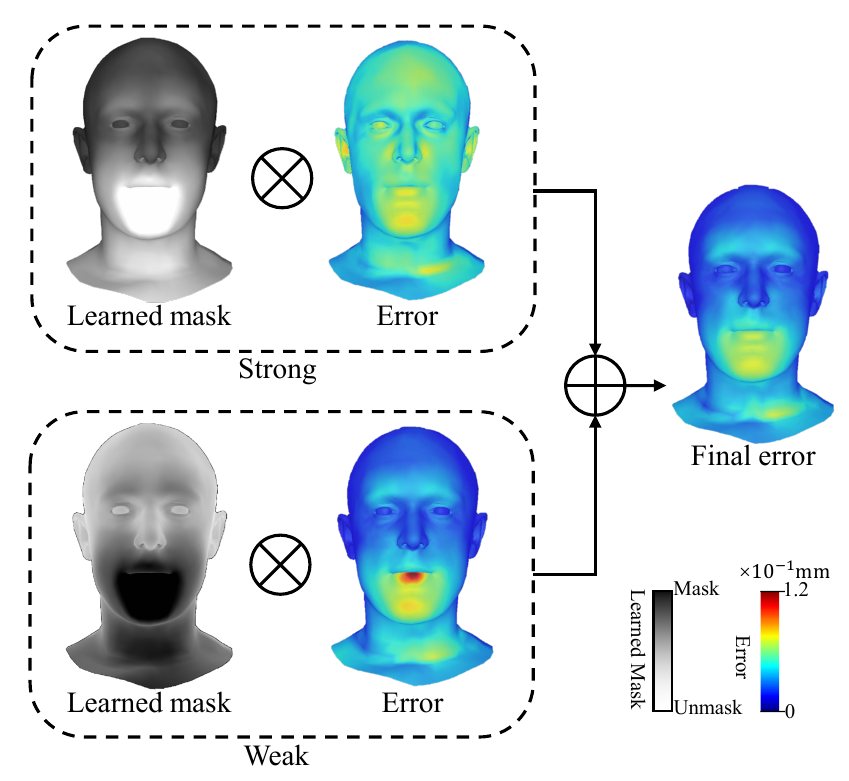}
    \caption{Visualization of the mean error of the synthetic sequences \( \mathbf{\tilde{Y}}^{S} \) (top: strong motion decoding branch ), \( \mathbf{\tilde{Y}}^{W} \) (bottom: weak motion decoding branch) and \(\mathbf{\hat{Y}} \) (right: final predicted sequence) out with the ground truth. }
    \label{merge_error}
\end{figure}

\subsubsection{Effectiveness of Dual-branch Decoder Based on FAI}
To investigate the effectiveness of the dual-branch decoding strategy and the priori intensity \(\mathbf{I_0}\) of facial activity, we first implement a modified model with a single branch to decode the entire facial animations, and then initialize \(\mathbf{m}_{t}\) with a randomly in a dual-branch decoder.

\noindent \textbf{Impact of the Dual-branch Decoder.} We implement a modified model with single branch to decode the entire facial animations. As shown in Table~\ref{ablation}, the metrics are unsatisfactory, especially the lip vertex error, which increases from 4.0858 to 4.5201. This is attributed to the difficulty of balancing intense, and high-frequency facial activities with mild, and low-frequency facial movements using a single branch. However, the dual-branch decoder provides an efficient method to help each branch focus on specific and similar intensity motion decoding, \textit{i.e.,} a specific branch performs professional tasks. The dual-branch decoding strategy effectively balances differences in facial movement intensity to produce accurate, natural facial movements. This result further confirms the need to consider differences in intensity of facial activity.

\noindent \textbf{Impact of the Mask Initialization Method.} A dual-branch decoding strategy based on the FAI can accommodate facial movements of different intensities, however the consistency of the two branches depends on the constraints of the mask. To investigate the impact of the mask initialization method, we randomly initialize the learnable mask and retrain a model. As seen in Table~\ref{ablation}, a randomly initialized mask diminishes the performance of this framework. Initializing the learnable mask using the motion prior \(\mathbf{I}_0\) is an effective way to prevent model performance degradation. Evidence suggests that how a learnable mask is initialised is critical for the performance of our framework.

\subsubsection{Impact of the Acoustic Feature Extractor}

The wavLM has shown a remarkable ability to extract speech representations in the field of speech recognition, emotion analysis. Then, wav2vec 2.0 is often employed as the speech feature extractor to generate facial animations\cite{fan2022faceformer,xing2023codetalker}. Therefore, we replace wavLM with wav2vec 2.0 as the acoustic feature extractor in CorrTalk and retrain the framework. The results are shown in Table~\ref{ablation}. Although the facial animation synthesized using wav2vec 2.0 has higher lip vertex errors and weak upper face dynamics, this results still exceed existing state-of-the-art methods. The phenomenon shows that our framework for wav2vec 2.0 is also effective for capturing the temporal correlation between speech and facial activities, and that wavLM has excellent capabilities to synthesize more accurate and natural facial motions for speech-driven 3D facial animation tasks.

\section{Conclusion}

In conclusion, we propose a novel driver framework, CorrTalk, to effectively capture the temporal correlation between HSF and facial activities of different intensities. The framework considers the differences in FAI and the heterogeneity of speech representations across different levels. The weighted hierarchical feature encoder provides a complementary and comprehensive speech representations to individually accommodate facial activities of varying intensity. The dual-branch decoding  based on the FAI synchronously produces strong and weak facial motions. Extensive experiments demonstrate that CorrTalk outperforms existing state-of-the-art methods in achieving accurate lip synchronization and generating realistic 3D facial animation. This work has promising applications in various fields, including entertainment, virtual reality, and human-computer interaction. However, the performance of CorrTalk is subject to limitations such as the training data quality and diversity. These limitations offer opportunities for future research to enhance the accuracy and generalization capabilities of CorrTalk.

\ifCLASSOPTIONcaptionsoff
  \newpage
\fi
\bibliographystyle{IEEEtran}
\bibliography{bibliography}

\vfill

\end{document}